\RequirePackage{amsmath}
\documentclass[runningheads]{llncs}
\makeatletter
\newcommand{\@chapapp}{\relax}%
\makeatother

\usepackage[title,toc,titletoc,header]{appendix}
\usepackage[T1]{fontenc}
\usepackage{booktabs} 
\usepackage{neuralnetwork}
\usepackage{lscape}
\usepackage{rotating}
\usepackage{longtable}
\usepackage{graphicx}
\usepackage[utf8]{inputenc} 
\usepackage[T1]{fontenc}    
\usepackage{hyperref}       
\usepackage{url}            
\usepackage{booktabs}       
\usepackage{amsfonts}       
\usepackage{nicefrac}       
\usepackage{microtype}      
\usepackage{xcolor}         

\usepackage{latexsym}
\usepackage{amssymb}
\usepackage{subcaption}
\usepackage{xmpmulti}
\usepackage{amsmath}
\usepackage{venndiagram}
\usepackage{tikz}
\usepackage{tikz-cd}
\usetikzlibrary{calc,cd,fit,shapes.geometric,tikzmark}
\tikzcdset{arrow style=tikz, diagrams={>={Classical TikZ Rightarrow[]}}}
\usetikzlibrary{calc,cd,fit,shapes.geometric,tikzmark}
\usepackage{algorithm}
\usepackage{algpseudocode}

\newcommand{\la}{\mathcal{L}}
\newcommand{\var}{\mathcal{V}}
\newcommand{\fa}{\mathcal{D}}

\begin{document}
\title{Semantic Objective Functions:\\
A distribution-aware method for adding logical constraints in deep learning}
\titlerunning{SOF:A method for adding logical constraints in deep learning}
%
\author{Miguel Angel Mendez-Lucero\inst{1}\orcidID{0000-0001-8349-8606} \and
Enrique Bojorquez Gallardo\inst{2}\orcidID{0009-0000-3321-7685} \and
Vaishak Belle\inst{3}\orcidID{0000-0001-5573-8465}}

\authorrunning{Miguel Angel Mendez-Lucero et al.}
%
\institute{School of Informatics, University of Edinburgh, Edinburgh, United Kingdom \\
\email{miguel.mendez@ed.ac.uk} \and
  Cohesive Solutions, Ciudad de Mexico, Mexico\\
\email{ebojorquez@cohesivesolutions.ai} \and
School of Informatics, University of Edinburgh, Edinburgh, United Kingdom \\
\email{vbelle@ed.ac.uk}}
\maketitle              
\begin{abstract}
Issues of safety, explainability, and efficiency are of increasing concern in learning systems deployed with hard and soft constraints. Symbolic Constrained Learning and Knowledge Distillation techniques have shown promising results in this area, by embedding and extracting knowledge, as well as providing logical constraints during neural network training. Although many frameworks exist to date, through an integration of  logic and information geometry, we provide a construction and theoretical framework for these tasks that generalize many approaches. We propose a loss-based method that embeds knowledge—enforces logical constraints—into a machine learning model that outputs probability distributions. This is done by constructing a  distribution from the external knowledge/logic formula, and constructing a loss function as a linear combination of the original loss function with the Fisher-Rao distance or Kullback-Leibler divergence to the constraint distribution. This construction includes logical constraints in the form of propositional formulas (Boolean variables), formulas of a first-order language with finite variables over a model with compact domain (categorical and continuous variables), and in general,likely applicable to any statistical model that was pretrained with semantic information. We evaluate our method on a variety of learning tasks, including classification tasks with logic constraints, transferring knowledge from logic formulas, and knowledge distillation from general distributions. 
\end{abstract}
\section{Introduction}\label{sec:Introduction}
Neuro symbolic artificial intelligence (NeSy) has emerged as a powerful tool for representing and reasoning about structured knowledge in deep learning \cite{survbelle}. Within this area, loss-based methods are a set of techniques that provide an integration of logic constraints into the learning process of neural architectures \cite{dllogicconst}. Logic constraints are logical formulas that represent knowledge about the problem domain. Such representations can be used to improve the accuracy, data, parametric efficiency, interpretability, and safety of deep learning models \cite{survbelle}. For example, in robotics, logic constraints can be used to represent safety constraints \cite{kwiatkowska2020safety,amodei2016concrete}. This information can help the model make more accurate predictions and provide interpretability, as we are embedding human expert knowledge into the network \cite{gunning2017explainable,rudin2018please}. Additionally, such expert knowledge reduces the problem space as the agent does not need to learn how to avoid harmful states and/or explore suboptimal policies, reducing the amount of data required to train a deep learning model \cite{multiplex_networks}. Broadly speaking, these neurosymbolic techniques solve the problem of obtaining distributions that \emph{satisfy the constraints} while solving a particular task. Our approach is different, we aim to \emph{learn the constraint} i.e. obtain a model that can potentially sample \emph{all} the instances of the constraint--for which satisfying the constraint becomes a consequence. Finally, this framework can also be used for Knowledge Distillation (KD) \cite{kdsurvey}, which consists in using an expert model with a complex architecture and transferring its information into a model with a simpler architecture, reducing size and complexity.

\subsection{Problem Description}\label{sec:problem}
Suppose we have a task that consists in training a statistical model to be able to describe a set of objects $X$. A description consists on assigning to each variable from a finite set $\var= \{x_1,..., x_n\}$ a value in a set $A$. Therefore, the possible descriptions are contained in the set $A^\var$.  The descriptive instances $a\in A^\var$ are the samples, $A^\var$ is the sample space, the variables the \emph{features} and the set of values $A$ is the \emph{domain}. Lets also assume that a single description may not capture completely what the object $x\in X$ is, maybe there are some descriptions that are equally good, or others that are better or worse. This can be accounted for if our statistical model assigns to each object $x\in X$ a probability distribution over $A^\var$. Asking the model what the object $x$ is, consists in sampling a description from this distribution, the probability of the sample shows how adequate it is as a description. It is the distribution that holds the information about the object---which is limited by the expressiveness of the sample space. Let us also assume that the model can only associate states from a family of distributions $\fa$. Therefore, our statistical model consists of a function,
\begin{equation}
    F: X\to \fa.
\label{eq:statistical_model}
\end{equation}
This model will be trained by minimizing a loss function $L: \fa\to\mathbb{R}^{\geq 0}$. 

What if there is external information that we want our model to learn as well? This extra information may come in the form of a formula that expresses the relationships between the variables. To exemplify this, if the variables represent features that can be true or false, then the domain can be seen as the set $\{0, 1\}$ (e.g. the Boolean space) and the constraint has the form of a propositional formula generated by a subset of the variables, such as $x_1\land \neg x_3$; if the domain is $\mathbb{R}$, then the formula may determine relationships expressed as inequalities that the outcomes of the variables must hold, such as $x_1^2 + x_2^2 + x_3^2 \leq 1$. These formulas act as constraints over the sample space, they represent knowledge that we want to embed into the system. Not only can we have extra information on the form of a specific region of the sample space, it can be a distribution over a region, allowing more complexity on the constraint. We will refer to the distribution that encodes this additional information as the \emph{constraint distribution}. The problem we want to solve now is: How can we train a statistical model for a given task in a way that it also learns this additional information? 

In this paper we provide a general framework for solving the problem  of transferring information from a given distribution that may come from a logic constraint or a pretrained neural network. Our contributions can be summarized as follows: we provide a canonical way to transform a constraint---in the form of propositional or First Order Logic (FOL) formulas of a language with finite variables---into a distribution, and then construct a loss function out of it. By minimizing over this loss function we can obtain a model that \emph{learns all possible instances of the constraint}, and as a consequence, \emph{learns the constraint}. We can also use this method to distill the information from a pretrained model allowing for more parameter efficient networks that also respect the logic constraints. 

The paper is divided into six sections. Section~\ref{sec:RelatedWork} contains related work on the problem of finding deep learning methods with logical constraints and KD. Section~\ref{sec:back_theory} presents the main concepts from information geometry and mathematical logic that will be used for the construction of the Semantic Objective Functions (SOFs). Section~ \ref{sec:sof} describes the construction of SOFs given a constraint distribution. Section~\ref{sec:experiments} has experiments that show that these methods solve the problem of learning the constraint in the case of propositional formulas, an experiment with the MNIST and Fashion MNIST databases, another using a FOL formula, and a KD problem. Section~\ref{sec:limitations} states the limitations of our approach, as well as some avenues for future work. Finally, Section~\ref{sec:conclusion} concludes the paper.
\section{Related Work}\label{sec:RelatedWork}

There are multiple approaches for enforcing logic constraints into the training of deep learning models \cite{survbelle,dllogicconst}. In this section, we will provide a general overview of the approaches that relate to our contributions and were used as basis of our work.

There are several techniques proposed to incorporate logical constraints into deep learning models. For propositional formulas there are semantic loss function \cite{semanticLossXu2018}, and LENSR \cite{emmb_log}. For FOL formulas there are: DL2 \cite{dl2}, for formulas of a relational language, MultiplexNet \cite{multiplex_networks} for formulas in disjunctive normal form and be quantifier-free linear arithmetic predicates, and  DeepProbLog \cite{deepproblog}, which encapsulates the outputs of the neural network in the form of neural predicates and learns a distribution to increase the probability of satisfaction of the constraints.

In addition to regularization methods, Knowledge Distillation (KD) offers another approach for transferring structured knowledge from logic rules to neural networks. Examples include methods proposed in \cite{hu2016harnessing} and \cite{RegDNKD}. These works build a rule-regularized neural network as a teacher and train a student network to mimic the teacher's predictions. A different approach is Concordia \cite{feldstein2023parallel}, an innovative system that combines probabilistic logical theories with deep neural networks for the teacher and student, respectively. 

Our framework differentiates itself from previous methods in certain key aspects. First, unlike \cite{hu2016harnessing} and \cite{RegDNKD}, it does not require the additional step of training the teacher network at each iteration. Second, compared to \cite{hu2016harnessing} and \cite{RegDNKD} and \cite{feldstein2023parallel}, it only requires a single loss function for training. This loss function can be used for training the network, regardless of whether the teacher is a logic constraint, a pre-trained network, or another kind of statistical model. This flexibility makes the loss function suitable for both semantic and deep learning KD scenarios. In the case of the loss-based methods, they build a regularizing loss function such that its value is zero whenever the model---or in the case of distributions, the support--- satisfies the formula---or is contained in the samples that satisfy the constraint. This makes sense,  as we do not want the regularizing term to penalize the distributions that satisfy the constraints. Given that the regularizer is a positive function, then all the elements that have zero value are local minimal, meaning that these approaches---as well as the other approaches---solve the problem of \emph{satisfying the constraint}. whereas our aim is to build a loss function that can \emph{learn the constraint} i.e. has as its \emph{unique minimal value} the uniform distribution over the set of models of the constraint. 

Another limitation is that their solution requires the constraint to be the same for all object $x\in X$, whereas in our approach we can have a constraint that depends on the object $x$. The only restrictions for being able to use our proposed methodology is that the constraint has to be a formula of a language generated by a \emph{finite} set of variables, and the set of instances that satisfy the formula has non-zero finite measure.

\section{Background Theory and Notation}\label{sec:back_theory}
The description is compressed for reasons of space, please refer to the Appendix \footnote{Appendix will be provided in a separate file}, also see \cite{lee,gallot,bell,enderton,cover,amari,amari2,Csiszar,watanabe} for a more comprehensive exposition.
\subsection{Information Geometry}\label{sec:Inf_Geom}
In this paper we will take $\fa$ to be a family of distributions over a space $A^\var$ which is parametrized by a chart $\theta\in\Omega\mapsto p(x;\theta)\in\fa$, where $\Omega$ is an open subset of $\mathbb{R}^m$. It has the structure of a convex Riemannian manifold where the metric is defined through the Fisher Information metric\cite{amari2}. The distance in the manifold derived by this metric is known as the Fisher-Rao distance, which is the one we will be using, and for distributions $p$ and $q$ it will be denoted as $d_\fa(p, q)$. Another important "measure" between probability distributions that we are going to use is the Kullback-Leibler Divergence\cite{cover}. This measure, denoted for distributions $p, q$ as $D_{KL}(p||q)$, can be interpreted as "the inefficiency of assuming that the distribution is $q$ when the true distribution is $p$"\cite{cover}, and is defined as,
$D(p||q) =\sum_{x\in A^\var}p(x) log\left(\frac{p(x)}{q(x)}\right) =\int_{x\in A^\var} p(x) log\left(\frac{p(x)}{q(x)}\right)dx,$ for the finite and continuous case respectively.

\subsection{Mathematical Logic}\label{sec:math_logic}

The logic constraints we will be using are expressed as formulas of a formal---propositional or FOL---language generated by a \emph{finite} set of variables $\var = \{x_1,..., x_n\}$. The models of a propositional language can be seen as the assignments of a truth value to each variable, i.e., $Mod(\la_\var) = 2^\var = \{s: \var\to 2\}$ where $2 = \{0,1\}$. On the other hand, a FOL formal language, $\la_\var^1$, is specified by a type $\tau$, which is a set of relations, functions and constant symbols. The terms are generated by the set of variables $\var$, and the formulas are recursively constructed by taking as atomic formulas the relations applied to terms, and then recursively constructing the rest with the logical operations $\land, \lor, \neg$, and quantifiers $\forall$ and $\exists$. A model $\mathfrak{A}$ of $\la_\var^1$ consists of a set $A$--the domain of the model--and an interpretation of each symbol in $\tau$ in $A$, and it will associate each formula $\varphi$ to a set $M_\varphi\subseteq A^n$. This determines a notion of satisfaction within the model that is recursively defined over the set of formulas--- $\mathfrak{A}\models \varphi(a_1, ..., a_n)$ if and only if $ (a_1,..., a_n)\in M_\varphi$\footnote{For more details see \cite{enderton} }.  For each formula $\varphi$, the size of the region that it constraints is denoted as $A_\varphi$--which is $|M_\varphi|$ in the propositional case or when $A$ is finite, and $\int_{M_\varphi}dx$ when $A=\mathbb{R}$. The uniform distribution associated to a formula $\varphi$, denoted as $\rho_\varphi$, is defined as,
\begin{equation}
      \rho_\varphi(a) = \begin{cases}
     1/A_\varphi &\text{ if }x\in M_\varphi\\
     0 &\text{ otherwise.}\\
    \end{cases}
    \label{eq:uniform}
\end{equation}
This distribution is only well-defined whenever $0< A_\varphi < \infty$, meaning that $M_\varphi$ has finite non-zero measure.

\section{Semantic Objective Functions}\label{sec:sof}

Gradient optimization algorithms are commonly used in machine learning to optimize the parameters of a model by minimizing a loss function \cite{Goodfellow-et-al-2016}. These algorithms work by iteratively updating the model parameters in the direction of the negative gradient of the loss function, which results in a sequence of model parameter values that minimize the loss function.

To address this issue, regularizers are often used to constrain the model's parameters during training, preventing them from becoming too large or too complex. They are added to the loss function as penalty terms, and their effect is to add a bias to the gradient update of the model parameters. Given a loss function $L$ and a regularizing term $L'$, the regularized loss function is a linear combination $\alpha L + \beta L'$, where $\alpha, \beta \in \mathbb{R}$. Our construction also includes the case in which the constraint depends on the object $x\in X$, where the training dataset are of the form $\{(X_i, \rho_i, \varphi_i)\}_{i\in I}$ or $\{(X_i, \varphi_i)\}_{i\in I}$ for the supervised and unsupervised cases respectively. In this section we will introduce the Semantic Objective Functions (SOFs) for each case and write down the explicit form in some important finite and continuous cases.

\subsection{Propositional and Finite domain constraints}

 Given a constraint distribution $\rho$ that is propositional or is interpreted in a model with finite domain, then the Fisher distance has a closed form\cite{boris} and can be used to construct the semantic regularizer as $L_\rho(f) = d_\fa (\rho, f).$ For these cases the family $\fa$ is the $n-1$-simplex $\{f: A^n\to [0,1] | \sum_{s\in 2^n}f(s) = 1\}$. For the propositional case $A=\{0,1\}$, and for the finite categorical case we are working on a model $\mathfrak{A}$ of a FOL language of type $\tau$ with domain $A= \{a_1,..., a_n\}$. To each formula $\varphi$, its associated constraint distribution is the uniform distribution $\rho_\varphi\in\fa$ defined in \ref{eq:uniform}. In this case the Fisher distance is defined as,
\begin{equation}
  d_\fa(p, q) = arccos\left( \sum_{s\in A^n} \sqrt{p(s)}\sqrt{q(s)} \right).  
  \label{eq:fisherprop}
\end{equation}
Therefore, the semantic regularizer can be defined as,
\begin{equation}
L_\varphi(f) = arccos\left(\sum_{s\in M_\varphi}\frac{\sqrt{f(s)}}{\sqrt{|M_\varphi|}}\right)
\label{eq:regprop}
\end{equation}

\subsubsection{Continuous domain}\label{external}

This is the case where the constraint is given as a formula of a FOL language of type $\tau$ in a model $\mathfrak{A}$ with continuous domain $A$. The terms are generated by the finite set $\var$, therefore the sample space is $A^n$. For this case we will use the Kullback-Leibler Divergence. There is a restriction on the formulas that we can apply this construction to. They have to be formulas $\varphi\in\la_\var^1$ such that $M_\varphi$ is of finite non-zero measure, so that $\rho_\varphi$, as defined in \ref{eq:uniform}, is well defined. The semantic regularizer associated to $\varphi$ is, $L_\varphi(f) = D_{KL}(\rho_\varphi || f).$
This function can be rewritten as a sum---or integral---over the set $M_\varphi$, which we can obtain through knowledge compilation techniques\cite{darwiche2002knowledge}.

\subsection{Extending Weighted Model Counting}\label{sec:vwmi}
As mentioned in Section~\ref{sec:back_theory}, there are limitations on some semantic regularization techniques that use a function that assigns weights to each atomic formula. This is the case in \cite{semanticLossXu2018}, where the logic constraint regularization is defined as logarithm the weighted model counting (WMC) \cite{chavira2008wmc} which is defined as:
\begin{equation}\label{eq:wmc}
    WMC(\varphi, \omega) = \sum_{M\models\varphi} \prod_{\psi\in Lit}\omega(\psi).
\end{equation}
Weighted model Counting defines a weight function $\omega: Lit\to\mathbb{R}^{\geq 0}$ over the literals, so that we can then calculate the probability of satisfaction of each models. Instead, from Equation~\ref{eq:wmc}, we can generalize the weight function $\omega$ to a weight function for the models, which is of the form $\omega:M_\varphi \to \mathbb{R}$ subject to $\sum_{x \in Mod(\la^0_\var)}\omega(x) = 1$. That is, $WMC(\varphi,\omega)= \sum_{x \in M_\varphi}\omega(x).$
Therefore, a natural extension of Equation~\ref{eq:wmc} is to extend it to FOL logic constraints with a finite amount of variables as,

\begin{equation}\label{eq:vwmi}
W(\varphi,\omega)= \int_{ M_\varphi}\omega(x_1, ..., x_n) dx_1...dx_n.
\end{equation}

\section{Experiments}\label{sec:experiments}

This section showcases the capabilities of the SOFs framework through focused experiments. We aim to demonstrate its advantages and potential use cases, rather than achieve state-of-the-art performance (left for future work). We employ propositional logic formulas in experiments from Sections~\ref{exp:1} and \ref{exp:2} for clear demonstration. Experiment in Section~\ref{exp:3} explores Knowledge Distillation over pretrained models. Finally, experiment 4 (Section~\ref{exp:2}) illustrates using SOFs with first-order logic (FOL) formulas with finite variables. For each experiment, we provide a high-level overview of the experiment setting. In-depth details on neural architectures, training, hyperparameter tuning, hardware, and software can be found in Appendix \footnote{Appendix will be provided in a separate file}.

\subsection{Constraint Learning}\label{exp:1}

In this experiment, we compare our SOFs to some of the related work techniques that can only guarantee to satisfy the formula. We use the following objective functions: KL-Divergence (KL-Div), Fisher-Rao distance (Fisher), Weighted Model Counting (WMC)\cite{chavira2008wmc}, Semantic Loss (Semloss)\cite{semanticLossXu2018}, and the $L^2$-norm\footnote{The $L^2$ norm is the well-known Euclidean distance, which can be defined for distributions $p$ and $q$ as $L^2(p,q)=\sqrt{\sum_{x}(p(x)-q(x))^2}$.}. For the case of WMC and Semloss we will use the generalization of WMC provided in Equation~\ref{eq:vwmi} rather than the definition of Equation~\ref{eq:wmc}.

Our experiments demonstrate that SOFs excel at learning logical constraints, going beyond just satisfying them. This is because SOFs aim to maximize information about the constraint, while other methods only minimize violation probability.

To analyze this behavior, we trained models on all  two-variable Boolean functions (15 satisfiable formulas)\footnote{The complete experiments with all formulas and configurations of supports can be found in the Appendix.}.In Table~\ref{tab:exp1} we focused on the  $\varphi =XOR$ formula with a target uniform constraint distribution $\rho_\varphi=[0,.5,.5,0]$. We varied initial distributions to explore different relationships with the target constraint distribution (fully contained, disjoint, partial overlap).

We chose the Fourier Signal Perceptron (FSP) \cite{signal_perceptron} for its effectiveness in learning Boolean functions, which aligns with our task of learning constraints. The FSP acts as the $\omega$ function from Equation~\ref{eq:vwmi} in our setup. We trained the model for 2,000 steps using stochastic gradient descent with a learning rate of 0.01. To evaluate how well the model learned the constraint, we compared the predicted distribution with the true constraint distribution using three distance measures: KL-Divergence, L2-norm, and Fisher-Rao distance.  There was consistency among the different measures, and given its information theoretic interpretation, the results presented in Table~\ref{tab:exp1} take the KL-Divergence as the final error.

\begin{table}[ht]
  \caption{Learning $\varphi =XOR=[0,1,1,0], \rho_\varphi=[0, 0.5, 0.5, 0]$ using different initial distributions.}
  \label{tab:exp1}
  \centering
  \begin{tabular}{llllllllllllllll}
    \toprule
    Init. Dist.(approx)    &[1 0 0  0]&[0 1 0 0]&[.5 0 0 .5]&[.33 .33 .33 0]&[.25 .25 .25 .25]\\
    \midrule
    $L^2$-Norm & 0.01201    & 2.74626   & 1.29408   & 0.00113   & 0.00130  \\
    FisherD.(ours)  & \textbf{8.821e-06}  & \textbf{1.782e-05} & \textbf{2.444e-06} & \textbf{1.889e-05} & \textbf{5.364e-06}\\
    KL-Div(ours)  & 0.00097    & 0.00046   & 0.00102   & 0.00078   & 0.00082  \\
    -WMC    & 0.82146    & 4.01373   & 0.52928   & 0.02228   & 0.00114  \\
    Sloss   &  3.48912   & 4.01371   & 0.38087   & 0.02220   & 0.00111  \\
    \bottomrule
  \end{tabular}
\end{table}

 As you can see, our SOFs are able to achieve a much lower error than the other methods. This is because our SOFs are able to learn the constraint, while the other methods are only able to satisfy the constraint. 

\subsection{Classification Tasks with logical Constraints.}\label{exp:2}

We evaluate the effectiveness of Semantic Objective Functions (SOFs) as a regularizer for image classification. We compare SOFs with other regularization techniques on two popular benchmark datasets: MNIST\cite{lecun2010mnist}, containing handwritten digits 0-9, and Fashion MNIST\cite{xiao2017fashion}, with images of 10 clothing categories. Both datasets use grayscale images. To enforce a constraint that each image belongs to exactly one class, we utilize a logical formula $\varphi = \bigvee_{i= 1}^n ((\bigwedge_{j\neq i, j=1}^n\neg x_j)\land x_i)$ representing one-hot encoding.

Our classification model is a simple four-layer Multilayer Perceptron (MLP) with a structure of [784,512,512,10] units per layer.  We employ ReLU activation functions in all hidden layers for efficient learning. In the final layer, however, we deviate from the typical SoftMax activation used in single-class classification. Instead, we adopt a sigmoid function. This aligns with the work of \cite{semanticLossXu2018}, where the network's output represents the satisfaction probability for each class belonging to the image. To ensure a fair comparison and maintain consistency with their approach, we calculate the probability  distribution over the models using the weighing function defined in \cite{semanticLossXu2018}, rather than the more general form presented in Equation~\ref{eq:vwmi}. We incorporated the semantic regularizers (Semloss, WMC, $L^2$-norm, Fisher-Rao distance and KL-Divergence) as an additional loss term in an MLP with MSE as the main loss. Experiments tested different regularization weights $\lambda\in\{1, 0.1,0.01, 0.001, 0.0001\}$ during batched training (128 images, Adam optimizer, learning rate 0.001, 10 epochs). We defer a wider method comparison to future work.
\begin{table}[ht]
 \caption{Learning Classification tasks with semantic regularizers}
  \label{tab:exp2}
  \centering
  \begin{tabular}{lllllll}
    \toprule
    \multicolumn{4}{c}{Fashion-MNIST} & \multicolumn{2}{c} {MNIST}\\ 
    \cmidrule(r){2-3}
    \cmidrule(r){4-5}
    Name     & $\lambda$  & Acc\% & $\lambda$ & Acc\% \\ 
    \midrule
    WMC & 0.01 & 87.74$\pm$ .64& 0.01 & 97.99 $\pm$ .27\\ 
    Sloss & 0.0001 & 87.81 $\pm$ .64&0.0001 & 97.78 $\pm$ .28\\ 
    Fisher & 0.1 & 88.23 $\pm$ .63&0.1 & \textbf{98.44} $\pm$ .24\\ 
    KL-Div & 0.01 & \textbf{88.24} $\pm$ .63&0.001 & 98.35$\pm$ .24\\ 
    $L^2$ Norm & 0.01 & 88.22 $\pm$ .63&0.1 & 98.27 $\pm$ .25\\
    NoReg  & 0 &75.00&0&97.60\\
    \bottomrule
  \end{tabular}
\end{table}
Table~\ref{tab:exp2} shows the results of training the MLP with same initial parameters, using different regularizers. We performed a grid search over the different $\lambda$'s and displayed the best results. Semantic regularizers provide an advantage of improving the accuracy by 1\% using the One-Hot-Encoding formula. While it is not shown on the table, the regularizer provides faster convergence than without regularization. Another important result is that the accuracy doesn't vary much for the different $\lambda$ in the case of the KL-Div, Fisher distance and $L^2$-norm, their results never went lower than 94\%. This was not the case with the other regularizers, both Semloss and WMC went lower than 10\% in both datasets when $\lambda$ is large enough. We believe this has to do with the fact that their functions have a lot of minimal elements, whereas the rest only have one.\footnote{For the complete results of the hyperparameter $\lambda$ grid search please check Appendix}  

\subsection{Constraint learning through Knowledge Distillation on classification tasks.}\label{exp:3}

In this experiment we want to demonstrate the KD capabilities of our SOFs as explained in Section~\ref{sec:back_theory}. The expert model we take is the one that was trained during the experiments \ref{exp:2} and showed best performance per dataset. It is important to notice that this model not only knows how to classify, but it also satisfies the constraint. We trained a smaller MLP with layers [784,256,256,10] to learn how to solve the MNIST and Fashion-MNIST in two ways: using the SOFs as a regularizer, as in Experiment ~\ref{exp:2}, or as the main loss function, as a process of KD. In the case of the regularizer, we tested for hyperparameters $\lambda\in\{0,0.0001,0.001,0.01,0.1,1.0\}.$ and displayed the best results for each SOF. For the case of KD, given that we take the SOF as the total loss function, the only information we use to learn the task is the one provided by the expert model. The labels are not used in training, just for measuring the accuracy. In both cases we take an adam optimizer, batch size of 128,learning rate of 0.001 and train for 10 epochs.
\begin{table}[ht]
  \caption{Transfer Learning : Regularizers vs Knowledge Distillation}
  \label{tab:exp3}
  \centering
  \begin{tabular}{lllllll}
    \toprule
    \multicolumn{5}{c}{Regularizer} & \multicolumn{2}{c}{Knowledge Distillation} \\
    \cmidrule(r){2-5}
    \cmidrule(r){6-7}
    \multicolumn{4}{c}{Fahion-MNIST} & \multicolumn{1}{c}{MNIST}&\multicolumn{1}{c}{Fahion-MNIST} & \multicolumn{1}{c}{MNIST} \\
    Name     & $\lambda$  & Acc\% & $\lambda$ & Acc\%& Acc\%& Acc\%\\
    \midrule
    KL-Div       & 0.001 &\textbf{87.94}$\pm$.63& 1.0 &\textbf{98.20}$\pm$.26 &\textbf{87.57}$\pm$.64&\textbf{98.12}$\pm$.26\\
    $L^2$-Norm      & 1.0 &87.75$\pm$.64& 0.1 &98.02$\pm$.27 &87.26$\pm$.65&98.11$\pm$.26\\
    Fisher.D.    & 1.0 &87.60$\pm$.63& 0.01 &98.01$\pm$.27 &87.38$\pm$.65&97.98$\pm$.27\\
    \bottomrule
  \end{tabular}
\end{table}
Table~\ref{tab:exp3} shows the results of this experiment. The results show a small reduction on the accuracy of the model compared against the expert models used in training. In the case when the loss function was used as a regularizer, the accuracy was reduced by $0.3\%$ and $.24\%$, for Fashion-MNIST and  MNIST respectively. Using the KD method the loss is $0.67\%$, $0.24\%$ for Fashion-MNIST and MNIST respectively. It is important to point out that this loss on accuracy comes with a reduction in the amount of required parameters, which is half of the expert model. Another observation is that these models still remain more accurate than some models---Semloss and WMC---that were trained on experiments 2, see table~ \ref{tab:exp2}. 
\subsection{Preliminary Results for First Order Logic formula}\label{exp:4}
This experiment uses Semantic Objective Functions to learn a probability density function (pdf) that closely matches assignments satisfying a finite-variable FOL formula. We approximate a uniform distribution over these assignments because, in continuous domains, we're limited to choosing from a specific family of pdfs. If the formula's true pdf doesn't belong to this family (multivariate normal distributions in this case), the learned parameters will only be an approximation. This lets us leverage the formulas defined in the external case from Section~\ref{external}.

To show that our framework is independent of the type of FOL formula, we will be using the following formula over real variables:
\begin{equation}\label{eq:folex1}
    \varphi(x_1, x_2)= (x_1^2+x_2^2\leq 1)\land \neg\exists z((z\neq 0)\land((x + z^2 = 0)\lor(y+z^2=0)))
\end{equation}

Like in Sections~\ref{sec:vwmi}, in order to define our learning method we first need to calculate the set of valid assignments  $M_{\varphi}$. For this, a common practice is to rely on SMT solvers or knowledge compilation techniques to compute the set of assignments in which the formula is valid \cite{darwiche2002knowledge}. For this experiment however, we can observe through a quick inspection that the formula  defines geometrically the set of all points  on the first quadrant of the unitary circle with center in $(0,0)$, its area is $\frac{\pi}{4}$. With the valid ranges defined above and a change of variables to polar coordinates, were $x= rcos\vartheta$ and $y= rsen\vartheta$, we can define a "semantic loss"-like function as \cite{semanticLossXu2018} for continuous domain:

\begin{equation}\label{eq:wmifolex1polar}
W(\varphi,\theta)=\int_0^1\int_0^{\pi/2} f(rcos(\vartheta),rsin(\vartheta);\theta)r d\vartheta dr.  
\end{equation}

where $f(x,y;\theta)$ will be replaced by the definition of the bi-variate normal distribution  and we will define our learnable parameter $\theta=\mu$. That is, our neural architecture will be a simple bivariate normal distribution and our learnable parameter will be the mean vector $\mu$. Now for our SOF, since we need to only integrate over the valid assignments, and the uniform distribution will always give the same value over those assignments of the formula we can replace the value $\rho_{\varphi}(x,y)$  to  $\frac{1}{\frac{\pi}{4}}=\frac{4}{\pi}$. We also change the variables to polar coordinates, obtaining the expression
\begin{equation}\label{eq:dklfoleq_sumpolar}
D_{kl}(\rho_{\varphi}||f_{\theta})=-\frac{4}{\pi}\int_0^1\int_0^{\frac{\pi}{2}}log(f_{\theta}( r\cos \vartheta ,r\sin \vartheta ))r d\vartheta dr 
\end{equation}


With Equation~\ref{eq:vwmi}, we can define a FOL version of the WMC and Semantic Loss which should maximize the probability of satisfaction of the formula. We conducted the experiment over 10 different seeds, over 2000 epochs and by using Adam and stochastic gradient descent optimizer. The results of the experiment can be found on Table~\ref{tab:sofexpfol1}, where we display the best results from a grid search of learning rate $\alpha\in\{1, 0.1,0.01, 0.001, 0.0001\}$. To measure the difference between the target and learned density function we used the Total Variation Distance which is defined as, $\delta (p,q)=\frac{1}{2} \int | p(x)-q(x) | dx$.


\begin{table}[ht]
\caption{Total Variation Distance for different optimizers and loss functions of $\varphi$}
\centering
\begin{tabular}{|c|c|c|c|}
\hline
Optimizer & Loss & Learning Rate $\alpha$ & Avg. Total Variation Distance \\
\hline
ADAM & W & 0.0001 & $0.37111863 \pm 0.03492265$ \\
ADAM & logW & 0.0001 & $0.36836797 \pm 0.043502506$ \\
ADAM & KLDiv & 0.01 & \textbf{0.2161428 $\pm$ 7.386059e-06} \\
SGD & W & 0.0001 & $0.39597395 \pm 0.049845863$ \\
SGD & logW & 0.0001 & $0.37211606 \pm 0.04976438$ \\
SGD & KLDiv & 0.001 & \textbf{0.212498 $\pm $0.0048501617} \\
\hline
\end{tabular}
\label{tab:sofexpfol1}
\end{table}
As expected our SOF was the one to attain the lowest error in comparison with the W and Wlog functions. It is important to mention that none of the loss functions could arrive to the correct solution. The reason is due to the fact that the bi-variate normal distribution doesn't have a shape that could realistically cover the area defined by the formula. 
\section{Limitations and Future Work}\label{sec:limitations}
This work has several limitations that motivate future research directions. First, we assume that for each formula  $\varphi$, the set $M_\varphi$ is readily available. This assumption necessitates solving the propositional satisfiability problem (SAT) or the counting satisfiability problem (\#SAT) for propositional logic, and solving systems of inequalities for general functions for first-order logic. As mentioned in Section~\ref{sec:RelatedWork}, other approaches share this assumption and leverage knowledge compilation techniques to address it \cite{darwiche2002knowledge,barrett2009handbook}. An important avenue for future work is to develop approximation methods for the set of satisfying assignments that circumvent the need for explicitly computing $M_\varphi$. Second, our current approach is limited to cases where $M_\varphi$ has non-zero finite measure. Third, the Fisher-Rao distance may not always have a closed-form solution, which can lead to computational challenges. Further research is needed to determine when the Fisher-Rao distance is preferable to the KL-divergence, considering their computational complexity and effectiveness in various settings. Fourth, a comprehensive evaluation comparing our method with existing approaches is necessary. Finally, it would be interesting to investigate the effectiveness of our method for a combination of SOFs obtained from a set of constraints $\{\rho_i\}_{i\in I}$ as well as further experimental analysis on constraint satisfaction using KD methods as in Section~\ref{exp:3}.

\section{Conclusion}\label{sec:conclusion}
In this paper, we proposed a framework and general construction of loss functions out of constraints and expert models, which can be used to transfer knowledge and enforce constraints into learned models. We divided the SOF constructions into two cases, internal and external, and for each case, we looked at closed forms for both finite and continuous domains. We ran experiments for classification tasks and both SOFs, KL-Divergence and Fisher-Rao distance, showed better accuracy than the other proposals \cite{semanticLossXu2018,chavira2008wmc}. We also illustrated the use of SOFs for finite variable first-order logic formulas showing promising results. Finally, we conducted KD experiments for two scenarios: when the set of satisfying assignments (SOFs) was used as a regularizer and when it served as the main loss function. These experiments are promising, demonstrating similar accuracy compared to their teacher models while requiring only half the number of parameters.

\begin{credits}
\subsubsection{\ackname} 
Vaishak Belle was supported by a Royal Society Research Fellowship.
Miguel Angel Mendez Lucero was supported by CONACYT Mexico.
\subsubsection{\discintname}
\end{credits}
%
%
%
\bibliographystyle{splncs04}
\bibliography{bibliografy}

\end{document}